\documentclass[preprint,12pt,3p]{elsarticle}
\usepackage[utf8]{inputenc}
\usepackage{multirow}
\usepackage{array}
\usepackage[lofdepth,lotdepth]{subfig}
\usepackage[ruled]{algorithm2e}
\usepackage{times,amsmath,amssymb}
\usepackage[T1]{fontenc}
\usepackage[polutonikogreek,english]{babel}
\usepackage{float}
\usepackage{appendix}
\usepackage{colortbl}
\usepackage{enumitem} % For itemize customization
\usepackage{hyperref}

\newenvironment{Sub-algorithm}[1][htb]
  {% Update algorithm name
   \begin{algorithm}[#1]%
  }{\end{algorithm}}
\usepackage{setspace}
\usepackage{amssymb}

\journal{ }

\begin{document}

\begin{frontmatter}

\title{Refining Myocardial Infarction Detection: A Novel Multi-Modal Composite Kernel Strategy in One-Class Classification\tnoteref{t1}}
\tnotetext[t1]{Accepted at 2024 IEEE International Conference on Image Processing.}

\address[label1]{Faculty of Information Technology and Communication Sciences, Tampere University, Tampere, Finland}
\address[label2]{VTT Technical Research Centre of Finland, Tampere, Finland}
\address[label3]{Haltian, Finland}
\address[label4]{Department of Electrical Engineering, Qatar University, Doha, Qatar}
\address[label5]{Hamad Medical Corporation, Doha, Qatar}

\author[label1]{Muhammad Uzair Zahid\corref{cor1}}
\cortext[cor1]{corresponding author}
\ead{muhammaduzair.zahid@tuni.fi}
\author[label1,label2]{Aysen Degerli}

\author[label1,label3]{Fahad Sohrab}

\author[label4]{ Serkan Kiranyaz}

\author[label5]{Tahir Hamid}

\author[label5]{Rashid Mazhar}

\author[label1]{and Moncef Gabbouj}

\begin{abstract}
Early detection of myocardial infarction (MI), a critical condition arising from coronary artery disease (CAD), is vital to prevent further myocardial damage. This study introduces a novel method for early MI detection using a one-class classification (OCC) algorithm in echocardiography. Our study overcomes the challenge of limited echocardiography data availability by adopting a novel approach based on Multi-modal Subspace Support Vector Data Description. The proposed technique involves a specialized MI detection framework employing multi-view echocardiography incorporating a composite kernel in the non-linear projection trick, fusing Gaussian and Laplacian sigmoid functions. Additionally, we enhance the update strategy of the projection matrices by adapting maximization for both or one of the modalities in the optimization process. Our method boosts MI detection capability by efficiently transforming features extracted from echocardiography data into an optimized lower-dimensional subspace. The OCC model trained specifically on target class instances from the comprehensive HMC-QU dataset that includes multiple echocardiography views indicates a marked improvement in MI detection accuracy. Our findings reveal that our proposed multi-view approach achieves a geometric mean of 71.24\%, signifying a substantial advancement in echocardiography-based MI diagnosis and offering more precise and efficient diagnostic tools.
\end{abstract}

\end{frontmatter}
\section{Introduction}
World Health Organization (WHO) recently emphasized the significant global health impact of Coronary Artery Disease (CAD), accounting for 16\% of worldwide deaths \cite{who2020topcausesofdeath}. Among the most severe outcomes of CAD is myocardial infarction (MI), characterized by irreversible damage of the myocardium, leading to a critical need for prompt detection \cite{reed2017acute}. MI diagnosis primarily involves symptom identification, biochemical markers, electrocardiography (ECG) readings, and imaging assessments. However, the initial symptoms, such as shortness of breath and upper body pain, may not always be immediately evident \cite{thygesen2012third}. Biochemical indicators like high sensitivity cardiac troponin (hs-cTn) also take time to manifest at levels indicative of MI \cite{macrae2006assessing,esmaeilzadeh2013role}. While ECG provides valuable insights, its diagnostic capabilities can be limited and often lag behind imaging methods \cite{esmaeilzadeh2013role}. Echocardiography, as a non-invasive imaging technique, is crucial in early MI detection by identifying Regional Wall Motion Abnormalities (RWMA) in the affected myocardium \cite{porter2018clinical}. Its accessibility and cost-effectiveness make it an advantageous option for early MI detection \cite{chatzizisis2013echocardiographic}.

While echocardiography is an important tool for diagnosing MI, its effectiveness faces several limitations. The subjective nature of evaluating RWMAs and the common occurrence of low-quality, noisy echocardiography recordings pose significant hurdles \cite{porter2018clinical,degerli2021early}. These challenges have spurred the increased reliance on computer-aided diagnosis algorithms. Previous studies have tested these algorithms with limited or non-diverse echocardiography datasets, raising concerns about their reliability and robustness, especially those based on deep learning techniques \cite{suhling2005myocardial,jamal2001noninvasive,chalana1996multiple,omar2018automated}. In response, one-class classification (OCC) models, which are trained only with samples from the positive class, have emerged as a promising alternative \cite{sohrab2020boosting,sohrab2018subspace}. However, the application of OCC models in echocardiography data analysis is still relatively unexplored, with few studies venturing into this domain \cite{gong2019fetal,loh2020p203}.

\begin{figure*}[t!]
    \centering
    \includegraphics[width=\linewidth]{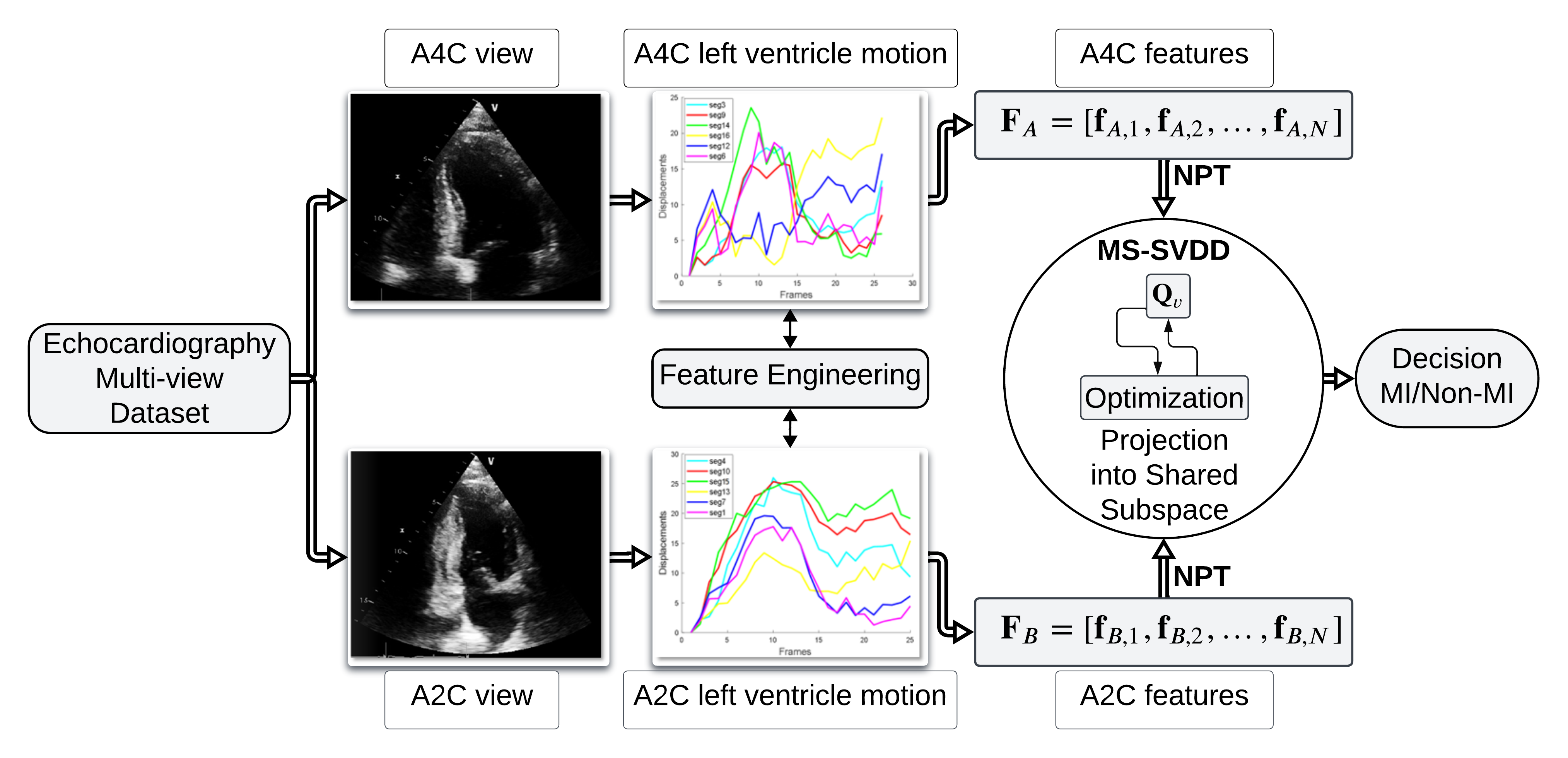}
    \caption{The proposed framework for early detection of MI with multi-modal one-class classiﬁcation over multi-view echocardiography. The workflow begins with image pairs representing the apical four-chamber (A4C) and apical two-chamber (A2C) views. Feature sets \( \mathbf{F}_A \) and \( \mathbf{F}_B \) extracted from the left ventricle motion characteristics undergo a transformation through NPT, followed by a projection into an optimized lower-dimensional shared space through the projection matrix \( \mathbf{Q} \). The process concludes with a multi-modal one-class classification that determines the presence or absence of MI.}
    \label{fig:framework}
\end{figure*}

In this study, we have developed a novel framework using OCC techniques for the early detection of MI using multi-view echocardiography. This framework focuses on extracting key features from apical 4-chamber (A4C) and apical 2-chamber (A2C) views by utilizing Active Polynomials (APs) to monitor left ventricle (LV) movement \cite{kiranyaz2020left}. A crucial aspect of the proposed method is the implementation of a composite kernel that combines Gaussian and Laplacian sigmoid kernels, designed to capture a more comprehensive representation of echocardiography data, thereby increasing the model's sensitivity to MI indicators. Moreover, we introduce an innovative strategy for optimizing the projection matrices, a central element in our OCC method. This optimization involves the combinations of gradient ascend and descent variations in the optimization process of data description and subspace learning to refine the model's performance. By fine-tuning the projection matrices, our model adapts more precisely to the distinctive MI features present in multi-view echocardiography. These advancements in composite kernel design and projection matrix optimization are extensively tested using the HMC-QU dataset. In addition to leveraging the advantages of multi-modal OCC, the proposed approach also explores the efficacy of both uni-modal and multi-modal algorithms in the early detection of MI through multi-view echocardiography.

The paper is organized in the following manner: Section \ref{sec:methods} presents the proposed framework for early MI detection with the details of the methodology and innovations introduced in the Multi-modal Subspace Support Vector Data Description (MS-SVDD) approach. Section \ref{sec:experiments} is dedicated to the experimental results and discussion including the analysis of the framework's performance on the HMC-QU\footnote{The benchmark HMC-QU is publicly shared at the repository \href{https://www.kaggle.com/datasets/aysendegerli/hmcqu-dataset}{https://www.kaggle.com/datasets/aysendegerli/hmcqu-dataset}.}  dataset. Finally, Section \ref{sec:conclusion} concludes the paper.

\section{Methodology}
\label{sec:methods}
As illustrated in Figure \ref{fig:framework}, the proposed OCC method is structured around the integration of three critical components: Multi-modal Subspace Support Vector Data Description, the composite kernel in non-linear projection trick (NPT), and adapting maximizing strategy for the projection matrix update. These elements work collectively to enhance the detection of MI through OCC using multi-view echocardiography. By combining these advanced techniques, our approach aims to improve the accuracy and efficiency of MI diagnosis.

\subsection{Feature Engineering and Data Preparation}
As in \cite{kiranyaz2020left}, the endocardial boundary of the left ventricle (LV) from A4C and A2C views of echocardiography recordings is extracted using Active Polynomials (APs). From the APs formed on each frame of the echocardiography recording, the extracted LV wall is tracked for myocardial motion analysis, where myocardial motion curves are obtained for each of the 12 myocardial segments. Then, the feature vectors are extracted from each myocardial segment motion curve in both A4C and A2C view echocardiography recordings over one cardiac cycle. Thus, A4C and A2C myocardial motion features are obtained for each myocardial segment in each echocardiography view, as depicted in Figure 1. Lastly, the extracted feature vectors are utilized for the detection of myocardial infarction with OCC techniques.

\subsection{Multi-modal Subspace Support Vector Data Description}
In the proposed framework, we used OCC for the detection of MI. Contrary to class-speciﬁc algorithms, the OCC models do not require information from negative samples during training. For the OCC model, we used the MS-SVDD as in \cite{sohrab2021multimodal} due to its feasibility with multi-view echocardiography data. MS-SVDD plays a pivotal role in our methodology, particularly in the context of multi-view echocardiography analysis. This approach focuses on mapping multi-view feature vectors into a lower-dimensional optimized feature space tailored for OCC.

The mathematical foundation of MS-SVDD is built around the concept of iteratively transforming the multi-modal (view) data from the original feature space to a new joint feature space and finding a joint compact description of data. Let the instances in each view $v$, $v=1,\dots,V$, are represented by $\mathbf{F}_v=[\mathbf{f}_{v,1},\mathbf{f}_{v,2},\dots\mathbf{f}_{v,N}]$, \(  \mathbf{f}_{v,i} \in \mathbb{R}^{D_v} \), where  \( D_v \) denotes the dimensionality of the feature space of view \( v \). A projection matrix \(  \mathbf{Q}_v \in \mathbb{R}^{d \times D_v} \) is learned to transform the feature vectors from the corresponding original \( D_v \)-dimensional space associated with view \( v \) into a lower \( d \)-dimensional shared subspace optimized for OCC. The total number of target class instances ($N$) is projected as follows:

\begin{equation}\label{eq:Y_i}
\mathbf{y}_{v,i} = \mathbf{Q}_v \mathbf{f}_{v,i},\forall v \in \{1,\dots,V\} \:\:, \forall i \in \{1,\dots,N\}.
\end{equation}

\noindent The objective of MS-SVDD is to find a compact hypersphere that encloses the target class data; the optimization function is defined as:
\begin{eqnarray}
    \min F(R, a) = R^2 + C \sum_{v=1}^{V} \sum_{i=1}^{N} \xi_{v,i} \nonumber\\
    \text{s.t.} \quad 
    || \mathbf{Q}_v  \mathbf{f}_{v,i} - a||^2 \leq R^2 + \xi_{v,i}, \nonumber \\
    \xi_{v,i} \geq 0, \quad \forall v \in \{1, \ldots, V\}, \forall i \in \{1, \ldots, N\},
\end{eqnarray}
\noindent where $R$ is the radius of the hypersphere, $a$ its center, $\xi_{v,i}$ are slack variables, and the term \( C \) controls the penalty for outliers in the training set. The Lagrangian function of MS-SVDD is as follows:
\begin{eqnarray}\label{Lang2}
L = \sum_{v=1}^{V}\sum_{i=1}^{N} \alpha _{v,i} \mathbf{y}_{v,i}^{\intercal}  \mathbf{y}_{v,i}-\sum_{v=1}^{V}\sum_{i=1}^{N}\sum_{n=1}^{V}\sum_{j=1}^{N} \alpha _{v,i} \mathbf{y}_{v,i}^{\intercal} \mathbf{y}_{n,j} \alpha _{n,j}.
\end{eqnarray}
where $\alpha$ is the Lagrange multiplier, and maximizing \eqref{Lang2} will give us $\alpha$ values for all training instances defining their position in the data description. It should be noted that optimizing \eqref{Lang2} for $\alpha$ corresponds to the traditional SVDD \cite{tax2004support} applied in the subspace \cite{tax2004support}. We update the projection matrix $ \mathbf{Q}_v$ iteratively as $ \mathbf{Q}_v \leftarrow  \mathbf{Q}_v - \eta \Delta L $, where $\Delta L$ is the gradient of the Lagrangian for the corresponding modality \( v \), and \( \eta \) is the learning rate. The gradient with respect to the projection matrix is defined as:

\begin{eqnarray}\label{eqforQ1}
\Delta L= 2\sum_{i=1}^{N} \alpha_{v,i} \mathbf{Q}_m \mathbf{f}_{v,i} \mathbf{f}_{v,i}^{\intercal}
- 2\sum_{i=1}^{N}\sum_{j=1}^{N}\sum_{n=1}^{V} \mathbf{Q}_n \mathbf{f}_{n,j} \mathbf{f}_{v,i}^{\intercal} \alpha_{v,i} \alpha_{n,j} + \beta\Delta \omega,
\end{eqnarray}

\noindent where $\omega$ is the regularization term incorporated into MS-SVDD, taking into account the covariance of data from different modalities in the shared subspace, the regularization term $\omega$ expresses the covariance and can take different forms depending on the instances taken into account during the regularization strategy. Its contribution is weighted by a hyper-parameter $\beta$. For a more in-depth understanding of various regularization techniques, readers can refer to \cite{sohrab2021multimodal}, where comprehensive details are provided. We maintain consistency in terminology and denote the regularization strategies using the same symbols outlined in the aforementioned reference.

This mathematical framework of MS-SVDD underpins our approach to optimizing the feature space for effective OCC in echocardiography. The incorporation of regularization techniques and the projection of features into a shared subspace are crucial aspects that enhance the robustness and accuracy of MI detection.

\subsection{MS-SVDD with Composite Kernel (MS-SVDD-CK)}
In our study, we have augmented the MS-SVDD by incorporating a composite kernel in the non-linear data description. The composite kernel is a crucial component in our methodology, combining the strengths of Gaussian and Laplacian sigmoid kernels. This combination improves the feature extraction capability of our model, which is crucial for effective myocardial infarction detection. The composite kernel is expressed as:

\begin{equation}
\mathbf{K}_{ij} = \gamma \exp\left(-\frac{||f_i - f_j||^2}{2\sigma^2}\right) + (1 - \gamma)  \tanh(\kappa  (f_i^{\intercal}f_j) + \theta),
\end{equation}
where $\gamma$ is the balancing parameter determining the weight given to each kernel type within the composite kernel, $\sigma$, is the scale parameter for the Gaussian kernel influencing the extent to which it captures variations in the data, and $\kappa$ and $\theta$ are unique parameters of the Laplacian sigmoid kernel. Consequently, $\kappa$ modulates the steepness of the sigmoid function, while $\theta$ shifts the function along the x-axis. These parameters allow the Laplacian component to adjust the behavior of the kernel function effectively. In this study, we set $\gamma=0.5$ to indicate equal importance to both the Gaussian and Laplacian sigmoid components. After obtaining the non-centered composite kernel matrix, we proceed with the Non-Linear Projection solution outlined in \cite{kwak2013nonlinear}. Subsequently, the composite kernel is centered as,
\begin{align}\label{centerK}
\mathbf{\Hat{K}} = (\mathbf{I}-  \frac{1}{N}\mathbf{1} \mathbf{1}^\intercal) \mathbf{K} ( \mathbf{I}- \frac{1}{N}\mathbf{1} \mathbf{1}^\intercal),
\end{align}
\noindent where $\mathbf{1} \in \mathbb{R}^N$ represents a vector with all elements set to one, and $\mathbf{I}\in\mathbb{R}^{N\times N}$ is an identity matrix. The centered kernel matrix $\mathbf{\Hat{K}}$ is decomposed by using eigendecomposition as follows:
\begin{align}\label{eigen}
\mathbf{\Hat{K}} = \mathbf{U}\mathbf{A}\mathbf{U}^\intercal, 
\end{align}
\noindent where $\mathbf{A}$ is a diagonal matrix containing the non-negative eigenvalues of the matrix $\mathbf{\Hat{K}}$ in its diagonal. The corresponding eigenvectors aligned with these eigenvalues are stored in the matrix $\mathbf{U}$ columns. The representation of data for non-linear data description is obtained as follows:
\begin{align}\label{nptdata}
{\Phi} = (\mathbf{A}^{\frac{1}{2}})^{+} \mathbf{U}^{+} {\mathbf{\Hat{K}}},
\end{align}
\noindent where $+$ in the superscript denotes the pseudo-inverse. After obtaining the data in $\Phi$ space, we apply all the steps for linear data description described in~\ref{algo}.

The proposed approach, merging both Gaussian and Laplacian sigmoid kernels, provides our Multi-modal Subspace Support Vector Data Description with Composite Kernel (MS-SVDD-CK) model with a robust and versatile tool for analyzing echocardiography data. The Gaussian component captures broad trends and general data characteristics, whereas the Laplacian sigmoid kernel focuses on detailed and local variations. This synergistic combination ensures a comprehensive analysis, enhancing the model's ability to detect myocardial infarction in echocardiography accurately.
\subsection{Optimization Techniques and Enhanced Adaptability}
We initialize the projection matrix $\mathbf{Q}$ for the corresponding v using Principal Component Analysis (PCA). The projection matrix is orthogonalized and normalized at every iteration so that $\mathbf{Q}\mathbf{Q}^\intercal=\mathbf{I}$. Our approach enhances the MS-SVDD model through the Symmetric Descent (SD) and Asymmetric Descent (AD) methods. These methods are designed to update the projection matrix $\mathbf{Q}_v$ more effectively, catering to the unique characteristics of each data modality. 

\textbf{Symmetric Descent Method:}
The SD method involves a consistent application of gradient operations across all modalities. It is denoted as SD- for gradient descent and SD+ for gradient ascent, applied uniformly across different data modalities. 

\textbf{Asymmetric Descent Method:}
The AD method applies divergent gradient operations to different modalities, employing descent for one and ascent for another. It allows for tailored updates according to the specific requirements of each modality. This method is represented by AD-+ and AD+-, indicating the direction of the gradient update for the first and second modalities, respectively. 

\textbf{Mathematical Formulation:}
The projection matrix $\mathbf{Q}$ for each modality \( v \) is updated using the gradient \( \Delta L( \mathbf{Q}^{(t)}) \) of the loss function at iteration \( t \) and the learning rate \( \eta \) as follows:

\begin{equation}\label{updateQ}
 \mathbf{Q}^{(t+1)} = 
\begin{cases} 
 \mathbf{Q}^{(t)} - \eta \Delta L( \mathbf{Q}^{(t)}) & \text{for SD-} \\
 \mathbf{Q}^{(t)} + \eta \Delta L( \mathbf{Q}^{(t)}) & \text{for SD+} \\
 \mathbf{Q}^{(t)} \mp \eta \Delta L( \mathbf{Q}^{(t)}) & \text{for AD-+ or AD+-} \\
\end{cases},
\end{equation}
\noindent where the $\mp$ symbol in the AD method indicates the application of negative gradient steps for the first modality and positive for the second in AD-+, and the reverse in AD+-. Algorithm \ref{algo} describes the overall MS-SVDD algorithm.

The SD and AD optimization strategies significantly enhance the MS-SVDD model. By fine-tuning the unique aspects of each data modality in the multi-view echocardiography dataset, these strategies ensure a more precise, robust, and efficient myocardial infarction detection model. The adaptability offered by SD and AD methods allows the MS-SVDD model to effectively address the complexities of multi-modal datasets, leading to improved accuracy and robustness. This approach is particularly effective for the early detection of myocardial infarction and has substantial potential to enhance anomaly detection and classification in real-world applications.

\begin{algorithm}[t!]
  \caption{The optimization process of the MS-SVDD algorithm.}\label{algo}
\SetAlgoLined
\SetKwInOut{Input}{Inputs}
\SetKwInOut{Output}{Outputs}
\Input{${\mathbf{F}}_v$ ($\Phi$ in NPT case) // Input data\\$\beta$ // Regularization parameter for $\omega$ \\$\eta$ // Learning rate parameter \\$d$, // Dimensionality of joint subspace \\$C$ // Regularization  parameter in SVDD \\$V$ // Total number of modalities (Views)} 
 \Output{$\textbf{Q}_v$ for each $v=1,...,V$ // Projection matrices\\$R$ // Radius of hypersphere \\$\mbox{\boldmath$\alpha$}$ // Defines the data description}  
              \vspace{3mm}   
\For{v=1:V}{
   \vspace{2mm}
   Initialize $\mathbf{Q}_v$ via PCA\;
 }
   \vspace{2mm}
   \For{$t=1:max\_iter$}{
       \vspace{2mm}
    For each $v$, map $\mathbf{F}_v$ to $\mathbf{Y}_v$ using Eq. (\ref{eq:Y_i})\;
          Form Y by combining all $\mathbf{Y}_v$'s\;
         \vspace{1mm}
        Solve SVDD in the subspace to obtain \boldmath$\alpha$ in Eq. \eqref{Lang2}\;
   \vspace{2mm}
  \For{v=1:V}{
   \vspace{2mm}
     Calculate $\Delta L$ using Eq. \eqref{eqforQ1}\;
    Update $\mathbf{Q}_v$ according to \eqref{updateQ}\;
        \vspace{2mm}
   Orthogonalize and normalize $\mathbf{Q}_v$ using QR decomposition;\\
  }
}
   For each $v$, compute $\mathbf{Y}_v$ using Eq. (\ref{eq:Y_i})\;
   Form $\mathbf{Y}$ by combining all $\mathbf{Y}_v$'s\;
   Solve SVDD to obtain the final data description\;
\end{algorithm}

\section{Experimental Evaluation}
\label{sec:experiments}
This section details our experimental setup and the results obtained over the benchmark HMC-QU dataset \cite{degerli2024early}. We compare the results with \cite{degerli2022early}, focusing on both multi-modal linear and non-linear approaches with novel optimization techniques and a composite kernel in the non-linear data description.
\subsection{Experimental Setup}

To evaluate the efficacy of our proposed framework, we conducted an extensive set of experiments on the HMC-QU dataset. This dataset comprises 260 echocardiography recordings, capturing both A4C and A2C views from a cohort of 130 individuals. The ground truths of the dataset are divided into two groups: 88 MI patients and 42 non-MI subjects. The implementation of all OCC models is carried out in MATLAB R2022b.

In the training phase of our OCC models, we designated MI and non-MI as the target classes, respectively. This allowed us to explore the model's performance in identifying both the presence and absence of MI in different settings. To comprehensively assess the effectiveness of our models, we employed several standard performance metrics, each offering a unique insight into the model's diagnostic capabilities: Sensitivity (Sen), indicating the proportion of true positives correctly identified; Specificity (Spe), reflecting the accurate identification of true negatives; Precision (Pre), measuring the rate of positive class predictions; F1-Score (F1), the harmonic mean of Sensitivity and Precision; Accuracy (Acc), the overall proportion of correct classifications across the dataset; and Geometric Mean (GM), representing the balance between Sensitivity and Specificity. Collectively, these metrics provide a comprehensive evaluation of our model's performance in identifying and differentiating MI and non-MI cases.

In our study, OCC models are extensively evaluated through a stratified 5-fold cross-validation (CV) scheme, allocating 80\% of the dataset for training and 20\% for testing. To optimize performance, we conducted an exhaustive search for the best hyperparameters over a stratified 10-fold CV scheme, focusing on maximizing GM during the training phase. The experimental assessment is conducted on the HMC-QU dataset, where we explore both linear and non-linear versions of the MS-SVDD models, along with the enhanced MS-SVDD with Composite Kernel, MS-SVDD-CK. The aim was to assess the efficacy of these models in myocardial infarction detection, employing various optimization strategies for improved accuracy. To ensure a fair comparison with previous studies, all parameters and settings were aligned with those cited in the referenced literature. The hyperparameters were configured as follows:

\begin{itemize}[leftmargin=2em, labelsep=1em]
\item Gaussian Kernel Scale ($\sigma$): The range $\{ 10^{-2}, 10^{-1}, 1, 10, 10^2, 10^3 \}$ is explored, allowing us to adjust the kernel's sensitivity to the spread of the data.
\item Laplacian Kernel Scale ($\kappa$): Set as $\kappa = \frac{1}{d}$, where $d$ is the dimensionality of the projected subspace. This scale parameter helps to control the kernel's focus on the local structure of the data.
\item Learning Rate ($\eta$): The range $\{ 10^{-4}, 10^{-3}, 10^{-2}, 10^{-1}, 1 \}$ is set for $\eta$. This parameter controls the rate of learning in the optimization process.

\item Regularization parameter ($\beta$): The values for $\beta$ were explored within $\{ 10^{-4}, 10^{-3}, 10^{-2}, 10^{-1}, 1, 10, 10^2, 10^3, 10^4 \}$, providing a range of options to balance the model complexity and control overfitting.

\item Penalty Parameter ($C$): We examined $C$ within the range $\{ 0.01, 0.05, 0.1, 0.2, 0.3, 0.4, 0.5, 0.6 \}$. This parameter determines the trade-off between maximizing the margin and minimizing classification errors.

\item Dimensionality ($d$): For multi-modal approaches, the projected dimensionality is set in the range $[1, 5]$, with increments of 1 at each step.

\item MS-SVDD Decision Strategies ($ds$): The MS-SVDD employs various decision strategies. Decision strategy 1 ($ds_1$) also called the AND gate assigns the target label if all modality representations classify to the target class; otherwise, it assigns the non-target label. Decision strategy 2 ($ds_2$) also called the OR gate assigns the target label if any modality representation classifies to the target class; otherwise, it assigns the non-target label. Decision strategy 3 ($ds_3$) bases the final classification on the representation from the first modality. Decision strategy 4 ($ds_4$) bases the overall decision on the label assigned to the representation from the second modality. The specifics of these strategies are detailed in the reference \cite{sohrab2021multimodal}.

\end{itemize}

\noindent The tuning of hyperparameters ensures that our approach is optimally configured to accurately and reliably detect myocardial infarction in echocardiography data; thus, enhancing the effectiveness of our diagnostic framework. 

\subsection{Results and Discussion}
A comprehensive study is conducted evaluating the performance of both MS-SVDD and MS-SVDD-CK models in OCC using the HMC-QU dataset, focusing on both MI and non-MI targets. The models are developed in both linear and non-linear versions, incorporating various optimization and decision strategies. The performance metrics are computed as the average of test sets of 5 folds in the HMC-QU dataset, as shown in Table \ref{tab:results}. In addition, Table \ref{tab:confusion_matrices} presents confusion matrices for optimal models based on the GM for MI and non-MI target classes.

\begin{table*}[t!]
\centering
\caption{Average myocardial infarction detection performance results (\%) computed over the test sets of each $5-$fold in HMC-QU dataset. Optimization strategies(OS): 'SD-' is represented by '\texttt{--}', 'SD+' by '++', 'AD-+' by '-+', and 'AD+-' by '+-',  NA (Not Applicable) designates when the OS and/or regularization is not applicable. }
\resizebox{\linewidth}{!}{
\begin{tabular}{cccccccccccccccccc}
& \multicolumn{8}{c}{\textbf{Target: MI}} &  & \multicolumn{8}{c}{\textbf{Target: non-MI}} \\
\cline{2-9} \cline{11-18}
& $OS$ & $r$ & Sen & Spe & Pre & F1 & Acc & GM &  & $OS$ & $r$ & Sen & Spe & Pre & F1 & Acc & GM  \\ 
\hline
\textit{Non-linear} \\
\hline

$\text{MS-SVDD-CK}_{ds_1}$&$-+$&$\omega_4$ & $70.45$ & $66.67$ & $81.58$ & $75.61$ & $69.23$ & $\bold{68.53}$ & &$++$& $\omega_0$& $73.81$ & $60.23$ & $46.97$ & $57.41$ & $64.62$ & $66.67$ \\

$\text{MS-SVDD-CK}_{ds_2}$&$-+$&$\omega_4$ & \cellcolor[gray]{.96}$56.82$ & \cellcolor[gray]{.96}$47.62$ & \cellcolor[gray]{.96}$69.44$ & \cellcolor[gray]{.96}$62.50 $ & \cellcolor[gray]{.96}$53.85$ & \cellcolor[gray]{.96}$52.02$ & &$++$&$\omega_2$ & \cellcolor[gray]{.96}$64.29$ & \cellcolor[gray]{.96}$64.77$ & \cellcolor[gray]{.96}$46.55$ & \cellcolor[gray]{.96}$54.00 $ & \cellcolor[gray]{.96}$64.62$ & \cellcolor[gray]{.96}$64.53$ \\

$\text{MS-SVDD-CK}_{ds_3}$&$+-$&$\omega_4$ & $56.82$ & $59.52$ & $74.63$ & $64.52$ & $5769$ & $58.16$ & & $+-$&$\omega_3$& $73.81$ & $55.68$ & $44.29$ & $55.36$ & $61.54$ & $64.11$ \\

$\text{MS-SVDD-CK}_{ds_4}$&$-+$&$\omega_5$ & \cellcolor[gray]{.96}$84.09$ & \cellcolor[gray]{.96}$42.86$ & \cellcolor[gray]{.96}$75.51$ & \cellcolor[gray]{.96}$79.57$ & \cellcolor[gray]{.96}$70.77$ & \cellcolor[gray]{.96}$60.03$ & &$++$&$\omega_6$ & \cellcolor[gray]{.96}$66.67$ & \cellcolor[gray]{.96}$76.14$ & \cellcolor[gray]{.96}$\bold{57.14}$ & \cellcolor[gray]{.96}$\bold{61.54}$ & \cellcolor[gray]{.96}$\bold{73.08}$ & \cellcolor[gray]{.96}$\bold{71.24} $ \\

$\text{MS-SVDD}_{ds_1}$&$-+$&$\omega_1$ & $67.05$ & $64.29$ & $79.73$ & $72.84$ & $66.15$ & $65.65$ & &$-+$& $\omega_1$& $71.43$ & $62.50$ & $47.62$ & $57.14$ & $65.38$ & $66.82$ \\

$\text{MS-SVDD}_{ds_2}$&$+-$&$\omega_3$ & \cellcolor[gray]{.96}$62.50$ & \cellcolor[gray]{.96}$42.86$ & \cellcolor[gray]{.96}$69.62$ & \cellcolor[gray]{.96}$65.87$ & \cellcolor[gray]{.96}$56.15$ & \cellcolor[gray]{.96}$51.75$ & &$++$&$\omega_3$ & \cellcolor[gray]{.96}$64.29$ & \cellcolor[gray]{.96}$63.64$ & \cellcolor[gray]{.96}$45.76$ & \cellcolor[gray]{.96}$53.47$ & \cellcolor[gray]{.96}$63.85$ & \cellcolor[gray]{.96}$63.96$ \\

$\text{MS-SVDD}_{ds_3}$&$-+$&$\omega_0$ & $77.27$ & $35.71$ & $71.58$ & $74.32$ & $63.85$ & $52.53$ & & $-+$&$\omega_2$& $69.05$ & $62.50$ & $46.77$ & $55.77$ & $64.62$ & $65.69$ \\

$\text{MS-SVDD}_{ds_4}$&$-+$&$\omega_1$ & \cellcolor[gray]{.96}$68.18$ & \cellcolor[gray]{.96}$61.90$ & \cellcolor[gray]{.96}$78.95$ & \cellcolor[gray]{.96}$73.17$ & \cellcolor[gray]{.96}$66.15$ & \cellcolor[gray]{.96}$64.97$ & &$-+$&$\omega_2$ & \cellcolor[gray]{.96}$71.43$ & \cellcolor[gray]{.96}$64.77$ & \cellcolor[gray]{.96}$49.18$ & \cellcolor[gray]{.96}$58.25$ & \cellcolor[gray]{.96}$66.92$ & \cellcolor[gray]{.96}$68.02$ \\

$\text{MS-SVDD}_{ds_1}$&$--$&$\omega_1$ & $70.45$ & $61.90$ & $79.49$ & $74.70$ & $67.69$ & $66.04$ & & $--$&$\omega_0$& $71.43$ & $53.41$ & $42.25$ & $53.10$ & $59.23$ & $61.77$ \\

$\text{MS-SVDD}_{ds_2}$&$--$&$\omega_2$ & \cellcolor[gray]{.96}$63.64$ & \cellcolor[gray]{.96}$42.86$ & \cellcolor[gray]{.96}$70.00$ & \cellcolor[gray]{.96}$66.67$ & \cellcolor[gray]{.96}$56.92$ & \cellcolor[gray]{.96}$52.23$ & &$--$&$\omega_0$ & \cellcolor[gray]{.96}$61.90$ & \cellcolor[gray]{.96}$73.86$ & \cellcolor[gray]{.96}$53.06$ & \cellcolor[gray]{.96}$57.14$ & \cellcolor[gray]{.96}$70.00$ & \cellcolor[gray]{.96}$67.62$ \\

$\text{MS-SVDD}_{ds_3}$&$--$&$\omega_2$ & $55.68$ & $59.52$ & $74.24$ & $63.64$ & $56.92$ & $57.57$ & &$--$&$\omega_0$ & $57.14$ & $57.95$ & $39.34$ & $46.60$ & $57.69$ & $57.54$\\

$\text{MS-SVDD}_{ds_4}$&$--$&$\omega_6$ & \cellcolor[gray]{.96}$39.77$ & \cellcolor[gray]{.96}$\bold{76.19}$ & \cellcolor[gray]{.96}$77.78$ & \cellcolor[gray]{.96}$52.63$ & \cellcolor[gray]{.96}$51.54$ & \cellcolor[gray]{.96}$55.05$ & &$--$&$\omega_2$ & \cellcolor[gray]{.96}$73.81$ & \cellcolor[gray]{.96}$67.05$ & \cellcolor[gray]{.96}$51.67$ & \cellcolor[gray]{.96}$60.78$ & \cellcolor[gray]{.96}$69.23$ & \cellcolor[gray]{.96}$70.35$ \\

$\text{ES-SVDD}$&NA&$\psi_0$ & $73.86$ & $38.10$ & $71.43$ & $72.63$ & $62.31$ & $53.05$ & &NA&$\psi_0$ & $69.05$ & $56.82$ & $43.28$ & $53.21$ & $60.77$ & $62.64$ \\

$\text{S-SVDD}$&NA&$\psi_2$ & \cellcolor[gray]{.96}$59.09$ & \cellcolor[gray]{.96}$54.76$ & \cellcolor[gray]{.96}$73.24$ & \cellcolor[gray]{.96}$65.41$ & \cellcolor[gray]{.96}$57.69$ & \cellcolor[gray]{.96}$56.88$ & &NA&$\psi_1$ & \cellcolor[gray]{.96}$54.76$ & \cellcolor[gray]{.96}$52.27$ & \cellcolor[gray]{.96}$35.38$ & \cellcolor[gray]{.96}$42.99$ & \cellcolor[gray]{.96}$53.08$ & \cellcolor[gray]{.96}$53.50$ \\

SVDD &NA&NA& $80.68$ & $38.10$ & $73.20$ & $76.76$ & $66.92$ & $55.44$ & &NA& NA& $69.05$ & $71.59$ & $53.70$ & $60.42$ & $70.77$ & $70.31$ \\

OC-SVM &NA&NA& $42.05$ & $71.43$ & $75.51$ & $54.01$ & $51.54$ & $54.81$ & &NA&NA & $35.71$ & $\bold{82.95}$ & $50.00$ & $41.67$ & $67.69$ & $54.43$ \\

\hline 
\textit{Linear}\\
\hline

$\text{MS-SVDD}_{ds_1}$&$++$ &$\omega_5$ & $72.73$ & $59.52$ & $79.01$ & $75.74$ & $68.46$ & $65.80$ &  &$-+$& $\omega_2$ & $71.43$ & $64.77$ & $49.18$ & $58.25$ & $66.92$ & $68.02$\\

$\text{MS-SVDD}_{ds_2}$&$-+$ & $\omega_5$& \cellcolor[gray]{.96}$60.23$ & \cellcolor[gray]{.96}$73.81$ & \cellcolor[gray]{.96}\cellcolor[gray]{.96}$82.81$ & \cellcolor[gray]{.96}$69.74$ & \cellcolor[gray]{.96}$64.62$ & \cellcolor[gray]{.96}$66.67$ & &$+-$&$\omega_2$ & \cellcolor[gray]{.96}$85.71$ & \cellcolor[gray]{.96}$47.73$ & \cellcolor[gray]{.96}$43.90$ & \cellcolor[gray]{.96}$58.06$ & \cellcolor[gray]{.96}$60.00$ & \cellcolor[gray]{.96}$63.96$ \\

$\text{MS-SVDD}_{ds_3}$&$+-$ &$\omega_5$ & $57.95$ & $76.19$ & $\bold{83.61}$ & $68.46$ & $63.85$ & $66.45$ & &$++$&$\omega_5$ & $66.67$ & $63.64$ & $46.67$ & $54.90$ & $64.62$ & $65.13$\\

$\text{MS-SVDD}_{ds_4}$&$-+$ &$\omega_5$ & \cellcolor[gray]{.96}$85.23$ & \cellcolor[gray]{.96}$47.62$ &\cellcolor[gray]{.96}$77.32$ & \cellcolor[gray]{.96}$\bold{81.08}$ & \cellcolor[gray]{.96}$\bold{73.08}$ & \cellcolor[gray]{.96}$63.71$ & &$++$& $\omega_4$& \cellcolor[gray]{.96}$73.81$ & \cellcolor[gray]{.96}$61.36$ & \cellcolor[gray]{.96}$47.69$ & \cellcolor[gray]{.96}$57.94$ & \cellcolor[gray]{.96}$65.38$ & \cellcolor[gray]{.96}$67.30$ \\

$\text{MS-SVDD}_{ds_1}$&$--$ &$\omega_5$ & $81.82$ & $47.62$ & $76.60$ & $79.12$ & $70.77$ & $62.42$ &  &$--$& $\omega_5$ & $73.81$ & $62.50$ & $48.44$ & $58.49$ & $66.15$ & $67.92$\\

$\text{MS-SVDD}_{ds_2}$&$--$ & $\omega_2$& \cellcolor[gray]{.96}$54.55$ & \cellcolor[gray]{.96}$59.52$ & \cellcolor[gray]{.96}\cellcolor[gray]{.96}$73.85$ & \cellcolor[gray]{.96}$62.75$ & \cellcolor[gray]{.96}$56.15$ & \cellcolor[gray]{.96}$56.98$ & &$--$&$\omega_2$ & \cellcolor[gray]{.96}$78.57$ & \cellcolor[gray]{.96}$36.36$ & \cellcolor[gray]{.96}$37.08$ & \cellcolor[gray]{.96}$50.38$ & \cellcolor[gray]{.96}$50.00$ & \cellcolor[gray]{.96}$53.45$ \\

$\text{MS-SVDD}_{ds_3}$&$--$ &$\omega_0$ & $67.05$ & $59.52$ & $77.63$ & $71.95$ & $64.62$ & $63.17$ & &$--$&$\omega_5$ & $73.81$ & $50.00$ & $41.33$ & $52.99$ & $57.69$ & $60.75$\\

$\text{MS-SVDD}_{ds_4}$&$--$ &$\omega_5$ & \cellcolor[gray]{.96}$85.23$ & \cellcolor[gray]{.96}$42.86$ &\cellcolor[gray]{.96}$75.76$ & \cellcolor[gray]{.96}$80.21$ & \cellcolor[gray]{.96}$71.54$ & \cellcolor[gray]{.96}$60.44$ & &$--$& $\omega_0$& \cellcolor[gray]{.96}$\bold{80.95}$ & \cellcolor[gray]{.96}$59.09$ & \cellcolor[gray]{.96}$48.57$ & \cellcolor[gray]{.96}$60.71$ & \cellcolor[gray]{.96}$66.15$ & \cellcolor[gray]{.96}$69.16$ \\

$\text{ES-SVDD}$&NA &$\psi_3$ & $82.95$ & $35.71$ & $73.00$ & $77.66$ & $67.69$ & $54.43$ & &NA&$\psi_3$ & $45.24$ & $67.05$ & $39.58$ & $42.22$ & $60.00$ & $55.08$ \\

$\text{S-SVDD}$&NA & $\psi_3$& \cellcolor[gray]{.96}$70.45$ & \cellcolor[gray]{.96}$45.24$ & \cellcolor[gray]{.96}$72.94$ & \cellcolor[gray]{.96}$71.68$ & \cellcolor[gray]{.96}$62.31$ & \cellcolor[gray]{.96}$56.45$ & &NA&$\psi_2$ & \cellcolor[gray]{.96}$50.00$ & \cellcolor[gray]{.96}$70.45$ & \cellcolor[gray]{.96}$44.68$ & \cellcolor[gray]{.96}$47.19$ & \cellcolor[gray]{.96}$63.85$ & \cellcolor[gray]{.96}$59.35$ \\

SVDD &NA&NA& $\bold{86.36}$ & $33.33$ & $73.08$ & $79.17$ & $69.23$ & $53.65$ & &NA& NA & $69.05$ & $69.32$ & $51.79$ & $59.18$ & $69.23$ & $69.18$ \\

OC-SVM &NA&NA & \cellcolor[gray]{.96}$44.32$ & \cellcolor[gray]{.96}$73.81$ & \cellcolor[gray]{.96}$78.00$ & \cellcolor[gray]{.96}$56.52$ & \cellcolor[gray]{.96}$53.85$ & \cellcolor[gray]{.96}$57.19$ & &NA& NA & \cellcolor[gray]{.96}$47.62$ & \cellcolor[gray]{.96}$81.82$ & \cellcolor[gray]{.96}$55.56$ & \cellcolor[gray]{.96}$51.28$ & \cellcolor[gray]{.96}$70.77$ & \cellcolor[gray]{.96}$62.42$ \\
\end{tabular}}
\label{tab:results}
\end{table*}

Moreover, we compare our results with uni-modal OCC algorithms, including the One-class Support Vector Machine (OC-SVM) \cite{scholkopfu2000sv}, the Support Vector Data Description (SVDD) \cite{tax2004support}, the Subspace SVDD (S-SVDD) \cite{sohrab2018subspace}, and the Ellipsoidal S-SVDD (ES-SVDD) \cite{sohrab2020ellipsoidal}. It is crucial to note that the last eight models for both linear and non-linear cases presented in Table \ref{tab:results} originate from our previous study \cite{degerli2022early}. In referenced uni-modal subspace OCC methods, S-SVDD and ES-SVDD), the regularization term is denoted by $\psi$ and represents the class variance in the projected space. Different regularization strategies ($\psi_0$-$\psi_3$) control the impact of individual samples in the regularization term. More details of the regularization strategies can be found in \cite{sohrab2023newton}. Using these benchmarks, a comparative analysis is conducted with the advancements constructed in the current study. The methodologies and optimization strategies employed in these models, as detailed in \cite{degerli2022early}, laid the foundation for the improvements realized in our latest research. By comparing the current study's outcomes with these established models, we can clearly illustrate the progress in myocardial infarction detection using echocardiography.

There is a significant improvement in MI detection performance with the MS-SVDD-CK models. The model $\text{MS-SVDD-CK}{ds_1}$, leveraging optimization AD-+ and composite kernel with regularization $\omega_4$, achieved a GM of 68.53\% for the MI target. This represents an approximate 3.77\% improvement over the previously best-performing model, $\text{MS-SVDD}{ds_2}$ with $\omega_5$ and SD- optimization, which had a GM of 66.04\%. For non-MI targets, the $\text{MS-SVDD-CK}{ds_4}$ model exhibited an even higher GM of 71.24\%, an enhancement of approximately 1.27\% compared to the previous model $\text{MS-SVDD}{ds_4}$ with a GM of 70.35\%. 

\begin{table}[t!]
\centering
\caption{Confusion matrices of optimal models for MI and Non-MI target classes.}
\label{tab:confusion_matrices}
\begin{tabular}{|c|c|c|}
\hline
\multicolumn{3}{|c|}{\textbf{$\text{MS-SVDD-CK}_{ds_1}$ (Target: MI)}} \\
\hline
 & \textbf{Predicted Non-MI} & \textbf{Predicted MI} \\
\hline
\textbf{Ground Truth Non-MI} & 28 & 14 \\
\hline
\textbf{Ground Truth MI} & 26 & 62 \\
\hline
\end{tabular}

\vspace{2mm} % Adds a little vertical space between the tables

\begin{tabular}{|c|c|c|}
\hline
\multicolumn{3}{|c|}{\textbf{$\text{MS-SVDD-CK}_{ds_4}$ (Target: non-MI)}} \\
\hline
 & \textbf{Predicted Non-MI} & \textbf{Predicted MI} \\
\hline
\textbf{Ground Truth Non-MI} & 28 & 14 \\
\hline
\textbf{Ground Truth MI} & 21 & 67 \\
\hline
\end{tabular}
\end{table}

The high precision of the linear $\text{MS-SVDD}_{ds_1}$ (AD+-) model suggests its potential utility in clinical settings where it is important to minimize false positives. There is, however, a trade-off implied by the lower sensitivity that may need to be addressed in future research. With a superior F1-Score, the linear $\text{MS-SVDD}_{ds_4}$ (AD-+) model is suitable for scenarios where both false positives and false negatives are of equal concern.

As a result of our study, it was found that non-linear models with composite kernels are particularly effective at balancing sensitivity and specificity, which is essential for reliable medical diagnosis. As indicated by the varied results across different configurations, optimization strategy, and regularization technique play a significant role in model performance.

\section{Conclusion}
\label{sec:conclusion}
This study presents a significant advancement in the detection of MI using multi-view echocardiography. Through the integration of a novel MS-SVDD-CK, we have enhanced the adaptability and effectiveness of OCC models. Incorporating Gaussian and Laplacian kernels with Symmetric and Asymmetric Descent methods has yielded notable improvements in model performance.
Based on the comprehensive experiments conducted over the HMC-QU dataset, we demonstrate a superior performance of our multi-modal approach, particularly its ability to adapt to different data modalities. This is evidenced by achieving the best geometric mean of 71.24\% for myocardial infarction detection. In addition to providing a robust framework for early MI diagnosis, this study also paves the way for future research in applying advanced machine learning techniques to cardiac health diagnostics. In the future, we will investigate using graph-embedded-based subspace learning methods \cite{sohrab2023graph} for myocardial infarction detection from multi-view echocardiography.

\bigskip
\noindent \textbf{Acknowledgments}

\noindent This work was supported by Research Council of Finland project AwCHA, and NSF-Business Finland project AMALIA. Foundation for Economic Education and Haltian's Carbon Handprint Research Program funded the work of Fahad Sohrab. Orion Research Foundation sr funded the work of Aysen Degerli.
\bibliographystyle{elsarticle-num}
\bibliography{ref}
\journal{ }
\end{document}